\documentclass[11pt,a4paper]{article}
\usepackage[hyperref]{emnlp-ijcnlp-2019}
\usepackage{times}
\usepackage{latexsym}
\usepackage{booktabs}

\usepackage{algorithm}
\usepackage{algpseudocode}
\usepackage{todonotes}
\usepackage{setspace}
\usepackage{url}
\usepackage{array}
\newcolumntype{L}{>{\arraybackslash}m{10cm}}
\usepackage{csquotes}

\aclfinalcopy 

\title{Neural data-to-text generation: \\ 
A comparison between pipeline and end-to-end architectures}

 \author{Thiago Castro Ferreira$^{1,2}$ \, Chris van der Lee$^{1}$ \, Emiel van Miltenburg$^{1}$ \, Emiel Krahmer$^{1}$ \\ 
         $^{1}$Tilburg center for Cognition and Communication (TiCC), Tilburg University, The Netherlands \\
         $^{2}$Federal University of Minas Gerais (UFMG), Brazil \\
         {\tt \{tcastrof,c.vdrlee\}@tilburguniversity.edu} \\
         {\tt \{c.w.j.vanmiltenburg,e.j.krahmer\}@tilburguniversity.edu}}

\date{}

\begin{document}
\maketitle
\begin{abstract}
  
  Traditionally, most data-to-text applications have been designed using a modular pipeline architecture, in which non-linguistic input data is converted into natural language through several intermediate transformations. By contrast, recent neural models for data-to-text generation have been proposed as end-to-end approaches, where the non-linguistic input is rendered in natural language with much less explicit intermediate representations in between. This study introduces a systematic comparison between neural pipeline and end-to-end data-to-text approaches for the generation of text from RDF triples. Both architectures were implemented making use of the encoder-decoder Gated-Recurrent Units (GRU) and Transformer, two state-of-the art deep learning methods. Automatic and human evaluations together with a qualitative analysis suggest that having explicit intermediate steps in the generation process results in better texts than the ones generated by end-to-end approaches. Moreover, the pipeline models generalize better to unseen inputs. Data and code are  publicly available.\footnote{\url{https://github.com/ThiagoCF05/DeepNLG/}}
  
\end{abstract}

\section{Introduction}

Data-to-text Natural Language Generation (NLG) is the computational process of generating meaningful and coherent natural language text to describe non-linguistic input data \cite{gatt2017}. Practical applications can be found in domains such as weather forecasts \citep{mei2016}, health care \citep{portet2009}, feedback for car drivers \citep{braun2018}, diet management \cite{mazzei2018}, election results \citep{leo2017} and sportscasting news \citep{chris2017}.

Traditionally, most of data-to-text applications have been designed in a modular fashion, in which the non-linguistic input data (be it, say, numerical weather information or game statistics) are converted into natural language (e.g., weather forecast, game report) through several explicit intermediate transformations. The most prominent example is the `traditional' pipeline architecture \citep{reiter2000} that performs tasks related to document planning, sentence planning and linguistic realization in sequence. 
Many of the traditional, rule-based NLG systems relied on modules because (a) these modules could be more easily reused across applications, and (b) because going directly from input to output using rules was simply too complex in general (see \citealt{gatt2017} for a discussion of different architectures). 

The emergence of neural methods changed this: provided there is enough training data, it does become possible to learn a direct mapping from input to output, as has also been shown in, for example, neural machine translation.
As a result, in NLG more recently, neural end-to-end data-to-text models have been proposed, which directly learn input-output mappings and rely much less on explicit intermediate representations \cite{wen2015,dusek2016,mei2016,lebret2016,gehrmann2018}.



However, the fact that neural end-to-end approaches are {\it possible}\/ does not necessarily entail that they are {\it better}\/ than (neural) pipeline models. On the one hand, cascading of errors is a known problem of pipeline models in general (an error in an early module will impact all later modules in the pipeline), which (almost by definition) does not apply to end-to-end models. On the other hand, it is also conceivable that developing dedicated neural modules for specific tasks leads to better performance on each of these successive tasks, and combining them might lead to better, and more reusable, output results. In fact, this has never been systematically studied, and this is the main goal of the current paper.

We present a systematic comparison between neural pipeline and end-to-end data-to-text approaches for the generation of output text from RDF input triples, relying on an augmented version of the WebNLG corpus \cite{claire2017b}. Using two state-of-the-art deep learning techniques, GRU \cite{choetal2014b} and Transformer \cite{vaswani2017}, we develop both a neural pipeline and an end-to-end architecture. The former, which also makes use of the NeuralREG approach \cite{ferreira2018}, tackles standard NLG tasks (discourse ordering, text structuring, lexicalization, referring expression generation and textual realization) in sequence, while the latter does not address these individual tasks, but directly tries to learn how to map RDF triples into corresponding output text.

Using a range of evaluation techniques, including both automatic and human measures, combined with a qualitative analysis, we provide answers to our two main research questions: (RQ1) How well do deep learning methods perform as individual modules in a data-to-text pipeline architecture? And (RQ2) How well does a neural pipeline architecture perform compared to a neural end-to-end one? Our results show that adding supervision during the data-to-text generation process, by distinguishing separate modules and combining them in a pipeline, leads to better results than full end-to-end approaches. Moreover, the pipeline architecture offers somewhat better generalization to unseen domains and compares favorably to the current state-of-the-art.

\section{Data}



The experiments presented in this work were conducted on the WebNLG corpus \cite{claire2017,claire2017b}, which consists of sets of $\langle$ Subject, Predicate, Object $\rangle$ RDF triples and their target texts. In comparison with other popular NLG benchmarks \cite{belz2011,novikova2017b,mille2018}, WebNLG is the most semantically varied corpus, consisting of 25,298 texts describing 9,674 sets of up to 7 RDF triples in 15 domains. Out of these domains, 5 are exclusively present in the test set, being unseen during the training and validation processes. \autoref{fig:modular} depicts an example of a set of 3 RDF triples and its related text.

To evaluate the intermediate stages between the triples and the target text, we use the augmented version of the WebNLG corpus \citep{ferreira2018b}, which provides gold-standard representations for traditional pipeline steps, such as discourse ordering (i.e., the order in which the source triples are verbalized in the target text), text structuring (i.e., the organization of the triples into paragraph and sentences), lexicalization (i.e., verbalization of the predicates) and referring expression generation (i.e., verbalization of the entities).

\begin{figure}
\footnotesize{
\begin{center}
\vspace{0.1cm}

\begin{tabular}{c c c}
A.C.\_Cesena   & \textbf{manager} & Massimo\_Drago \\
Massimo\_Drago & \textbf{club} & S.S.D.\_Potenza\_Calcio \\
Massimo\_Drago & \textbf{club} & Calcio\_Catania  \\
&  &   \\
 & $\downarrow$ &   \\
\end{tabular}
\end{center}
\vspace{0.1cm}

Massimo Drago played for the club SSD Potenza Calcio and his own club was Calcio Catania. He is currently managing AC Cesena.
\vspace{0.1cm}
\caption{Example of a set of triples (top) and the corresponding text (bottom).}
\label{fig:modular}
}
\end{figure}

\section{Pipeline Architecture}

Based on \citet{reiter2000}, we propose a pipeline architecture which converts a set of RDF triples into text in 5 sequential steps.

\subsection{Discourse Ordering}

Originally designed to be performed when the document is planned, Discourse Ordering is the process of determining the order in which the communicative goals should be verbalized in the target text. In our case, the communicative goals are the RDF triples received as input by the model.

Given a set of linearized triples, this step determines the order in which they should be verbalized. For example, given the triple set in \autoref{fig:modular} in the linearized format:


\vspace{0.3cm}
\begin{displayquote}
    \footnotesize
    \vspace{0.1cm}
    \noindent \texttt{<TRIPLE>} A.C.\_Cesena \textbf{manager} Massimo\_Drago \texttt{</TRIPLE>} \texttt{<TRIPLE>} Massimo\_Drago \textbf{club} S.S.D.\_Potenza\_Calcio \texttt{</TRIPLE>} \texttt{<TRIPLE>} Massimo\_Drago \textbf{club} Calcio\_Catania \texttt{</TRIPLE>}
    \vspace{0.1cm}
\end{displayquote}
\vspace{0.2cm}

Our discourse ordering model would ideally return the set \texttt{club club manager}, which later is used to retrieve the input triples on the predicted order. In case of triples with the same predicates, as \texttt{club}, our implementation will randomly retrieve the triples.

\subsection{Text Structuring}
Text Structuring is the step which intends to organize the ordered triples into paragraphs and sentences. Since the WebNLG corpus only contains single-paragraph texts, this step will be only evaluated on sentence planning, being closer to the Aggregation task of the original architecture \cite{reiter2000}. However, it can be easily extended to predict paragraph structuring in multi-paragraph datasets.

Given a linearized set of ordered triples, this step works by generating the predicates segmented by sentences based on the tokens \texttt{<SNT>} and \texttt{</SNT>}. For example, given the ordered triple set in \autoref{fig:modular} in the same linearized format as in Discourse Ordering, the module would generate \texttt{<SNT> club club </SNT> <SNT> manager </SNT>}, where predicates are replaced by the proper triples for the next step.

\subsection{Lexicalization}
Lexicalization involves finding the proper phrases and words to express the content to be included in each sentence \cite{reiter2000}. In this study, given a linearized ordered set of triples segmented by sentences, the Lexicalization step aims to predict a template which verbalizes the predicates of the triples. For our example based on \autoref{fig:modular}, given the ordered triple set segmented by sentences in the following format:

\vspace{0.3cm}
\begin{displayquote}
    \footnotesize
    \vspace{0.1cm}
    \noindent \texttt{<SNT>} \texttt{<TRIPLE>} Massimo\_Drago \textbf{club} S.S.D.\_Potenza\_Calcio \texttt{</TRIPLE>} \texttt{<TRIPLE>} Massimo\_Drago \textbf{club} Calcio\_Catania \texttt{</TRIPLE>} \texttt{</SNT>} \texttt{<SNT>} \texttt{<TRIPLE>} A.C.\_Cesena \textbf{manager} Massimo\_Drago \texttt{</TRIPLE>} \texttt{</SNT>}

    \vspace{0.1cm}
\end{displayquote}
\vspace{0.2cm}

This step would ideally return a template like:

\vspace{0.3cm}
\begin{displayquote}
    \footnotesize
    \vspace{0.1cm}
    \noindent \textbf{ENTITY-1} VP[\textit{aspect=simple, tense=past, voice=active, person=null, number=null}] play for DT[\textit{form=defined}] the club \textbf{ENTITY-2} and \textbf{ENTITY-1} own club VP[\textit{aspect=simple, tense=past, voice=active, person=null, number=singular}] be \textbf{ENTITY-3} . \textbf{ENTITY-1} VP[\textit{aspect=simple, tense=present, voice=active, person=3rd, number=singular}] be currently VP[\textit{aspect=progressive, tense=present, voice=active, person=null, number=null}] manage \textbf{ENTITY-4} .
    \vspace{0.1cm}
\end{displayquote}
\vspace{0.1cm}

The template format not only selects the proper phrases and words to verbalize the predicates, but also does indications for the further steps. The general tags \texttt{ENTITY-[0-9]} indicates where references should be realized. The number in an entity tag indicates the entity to be realized based on its occurrence in the ordered triple set. For instance, \texttt{ENTITY-3} refers to the entity \texttt{Calcio\_Catania}, the third mentioned entity in the ordered triple set.

Information for the further textual realization step is stored in the tags \texttt{VP}, which contains the aspect, mood, tense, voice and number of the subsequent lemmatized verb, and \texttt{DT}, which depicts the form of the subsequent lemmatized determiner.\footnote{Both kind of tags with their respective information are treated as a single token.}


\subsection{Referring Expression Generation}

Referring Expression Generation (REG) is the pipeline task responsible for generating the references to the entities of the discourse \cite{krahmer2012}. As previously explained, the template created in the previous step depicts where and to which entities such references should be generated. Given our example based in \autoref{fig:modular}, the result of the REG step for the template predicted in the previous step would be:

\vspace{0.3cm}
\begin{displayquote}
    \footnotesize
    \vspace{0.1cm}
    \noindent \textbf{Massimo Drago} \texttt{VP[...]} play for \texttt{DT[...]} the club \textbf{SSD Potenza Calcio} and \textbf{his} own club \texttt{VP[...]} be \textbf{Calcio Catania} . \textbf{He} \texttt{VP[...]} be currently \texttt{VP[...]} manage \textbf{AC Cesena} .
    \vspace{0.1cm}
\end{displayquote}
\vspace{0.1cm}

To perform the task, we used the NeuralREG algorithm \cite{ferreira2018}. Given a reference to be realized, this algorithm works by encoding the template before (pre-context) and after (post-contex) the reference using two different Bidirectional LSTMs \cite{hochreiter1997}. Attention vectors are then computed for both vectors and concatenated together with the embedding of the entity. Finally, this representation is decoded into the referring expression to the proper entity in the given context.\footnote{NeuralREG works with the Wikipedia representation of the entities (e.g., \texttt{Massimo\_Drago}) in the templates instead of general tags (e.g., \texttt{ENTITY-1}).}

\subsection{Textual Realization}

Textual Realization aims to perform the last steps of converting the non-linguistic data into text. In our pipeline architecture this includes setting the verbs (e.g., \textit{VP[aspect=simple, tense=past, voice=passive, person=3rd, number=singular] locate $\rightarrow$ was located}) and determiners (\textit{DT[form=undefined] a American national $\rightarrow$ an American national}) to their right formats. Both verbs and determiners are solved in a rule-based strategy, where the implications are extracted from the training set. This step will not be individually evaluated as the other ones.

\section{End-to-End Architecture}

Our end-to-end architecture is similar to novel data-to-text models \cite{wen2015,dusek2016,mei2016,lebret2016,gehrmann2018}, which aim to convert a non-linguistic input into natural language without explicit intermediate representations, making use of Neural Machine Translation techniques. In this study, our end-to-end architecture intends to directly convert an unordered (linearized) set of RDF triples into text.

\section{Models Set-Up}


Both pipeline steps and the end-to-end architecture were modelled using two deep learning encoder-decoder approaches: Gated-Recurrent Units (GRU; \citealt{choetal2014b}) and Transformer \cite{vaswani2017}. These models differ in the way they encode their input. GRUs encode the input data by going over the tokens one-by-one, while Transformers (which do not have a recurrent structure) may encode the entire source sequence as a whole, using position embeddings to keep track of the order. We are particularly interested in the capacity of such approaches to learn order and structure in the process of text generation. The model settings are explained in the supplementary materials.

\section{Experiment 1: Learning the pipeline}

Most of the data-to-text pipeline applications have their steps implemented using rule-based or statistical data-driven models. However, these techniques have shown to be outperformed by deep neural networks in other Computational Linguistics subfields and in particular pipeline steps like Referring Expression Generation. NeuralREG \cite{ferreira2018}, for instance, outperforms other techniques in generating references and co-references along a single-paragraph text. 
Given this context, our first experiment intends to analyze how well deep learning methods perform particular steps of the pipeline architecture, like Discourse Ordering, Text Structuring, Lexicalization and Referring Expression Generation, in comparison with simpler data-driven baselines.

\subsection{Data} 
We used version 1.5 of the augmented WebNLG corpus \cite{ferreira2018b}\footnote{\url{https://github.com/ThiagoCF05/webnlg}} to evaluate the steps of our pipeline approach. Based on its intermediate representations, we extracted gold-standards to train and evaluate the different steps.

\paragraph{Discourse Ordering} We used pairs of RDF triple sets and their ordered versions to evaluate our Discourse Ordering approaches. For the cases in the training set where a triple set was verbalized in more than one order, we added one entry per verbalization taking the proper order as the target. To make sure the source set followed a pattern, we ordered the input according to the alphabetic order of its predicates, followed by the alphabetical order of its subjects and objects in case of similar predicates. In total, our Discourse Ordering data consists of 13,757, 1,730 and 3,839 ordered triple sets for 5,152, 644 and 1,408 training, development and test input triple sets, respectively. 

\paragraph{Text Structuring} 14,010, 1,752 and 3,955 structured triple sets were extracted for 10,281, 1,278 and 2,774 training, development and test ordered triple sets, respectively.

\paragraph{Lexicalization} 18,295, 2,288 and 5,012 lexicalization templates were used for 12,814, 1,601 and 3,463 training, development and test structured triple sets, respectively.

\paragraph{Referring Expression Generation} To evaluate the performance of the REG models, we extracted 67,144, 8,294 and 19,210 reference instances from training, development and test part of the corpus. Each instance consists of the cased tokenized referring expression, the identifier of the target entity and the uncased tokenized pre- and post-contexts.

\subsection{Metrics}

Discourse Ordering and Text Structuring approaches were evaluated based on their accuracy to predict one of the gold-standards given the input (many of the RDF triple sets in the corpus were verbalized in more than one order and structure). Referring Expression Generation approaches were also evaluated based on their accuracy to predict the uncased tokenized gold-standard referring expressions. Lexicalization was evaluated based on the BLEU score of the predicted templates in their uncased tokenized form.

\subsection{Baselines}

We proposed \textit{random} and \textit{majority} baselines for the steps of Discourse Ordering, Text Structuring and Lexicalization. In comparison with NeuralREG, we used the \textit{OnlyNames} baseline, also introduced in \citet{ferreira2018}. 

\paragraph{Discourse Ordering} The random baseline returns the triple set in a random order, whereas the majority one returns the most frequent order of the input predicates in the training set. For unseen sets of predicates, the \textit{majority} model returns the triple set in the same order as the input.

    
        

\paragraph{Text Structuring} The random baseline for this step chooses a random split of triples in sentences, inserting the tags \texttt{<SNT>} and \texttt{</SNT>} in aleatory positions among them. The majority baseline returns the most frequent sentence intervals in the training set based on the input predicates. In case of an unseen set, the model looks for sentence intervals in subsets of the input.

\begin{algorithm}
\footnotesize{
\begin{algorithmic}[1]
    \Require struct, model 
    \State start, end $\gets$ 0, $|$struct$|$
    \State template $\gets \emptyset$
    
    \While{start $< |$struct$|$}
        \State snts $\gets$ struct[start,end)
        
        \If{snts $\in$ model}
            \State template $\gets$ template $\cup$ model[snts]
            \State start $\gets$ end
            \State end $\gets |$struct$|$
        \Else
            \State end $\gets$ end $- 1$
            
            \If{start $=$ end}
                \State start $\gets$ start + 1
                \State end $\gets |$struct$|$
            \EndIf
        \EndIf
    \EndWhile
    \State \Return template
\end{algorithmic}
}
\caption{Lexicalization Pseudocode}
\label{alg:lex}
\end{algorithm}

\paragraph{Lexicalization} Algorithm \ref{alg:lex} depicts our baseline approach for Lexicalization. As in Text Structuring, given a set of input triples structured in sentences, the random and majority models return a random and the most frequent template that describes the input predicates, respectively (line 6). If the set of predicates is unseen, the model returns a template that describes a subset of the input.

\paragraph{Referring Expression Generation} We used \textit{OnlyNames}, a baseline introduced in \citet{ferreira2018}, in contrast to NeuralREG. Given an entity to be referred to, this model returns the entity Wikipedia identifier with underscores replaced by spaces (\texttt{Massimo\_Drago} $\rightarrow$ \texttt{Massimo Drago}).

\begin{table}
\centering
\footnotesize{
\begin{tabular}{l c c c}
\toprule
                         & \textbf{All} & \textbf{Seen} & \textbf{Unseen} \\
\midrule
\multicolumn{4}{c}{\bf Discourse Ordering}\\
\midrule
Random      & 0.31 & 0.29 & 0.35 \\
Majority    & 0.48 & 0.51 & 0.44 \\
GRU         & 0.35 & 0.56 & 0.10 \\
Transformer & 0.34 & 0.56 & 0.09 \\
\midrule
\multicolumn{4}{c}{\bf Text Structuring}\\
\midrule
Random      & 0.29 & 0.29 & 0.30 \\
Majority    & 0.27 & 0.45 & 0.06 \\
GRU         & 0.39 & 0.63 & 0.13 \\
Transformer & 0.36 & 0.59 & 0.12 \\
\midrule
\multicolumn{4}{c}{\bf Lexicalization}\\
\midrule
Random      & 39.49 & 40.46 & 33.79 \\
Majority    & 44.82 & 45.65 & 39.43 \\
GRU         & 37.43 & 49.26 & 23.63 \\
Transformer & 38.12 & 48.14 & 24.15 \\
\midrule
\multicolumn{4}{c}{\bf Referring Expression Generation}\\
\midrule
\textit{OnlyNames} & 0.51 & 0.53 & 0.50 \\
NeuralREG          & 0.39 & 0.70 & 0.07 \\
\bottomrule
\end{tabular}
}
\caption{Accuracy of Discourse Ordering, Text Structuring and Referring Expression models, as well as BLEU score of Lexicalization approaches.}
\label{table:results1}
\end{table}

\subsection{Results} 

\autoref{table:results1} shows the results for our models for each of the 4 evaluated pipeline steps. In general, the results show a clear pattern in all of these steps: both neural models (GRU and Transformer) introduced higher results on domains seen during training, but their performance drops substantially on unseen domains in comparison with the baselines (Random and Majority). The only exception is found in Text Structuring, where the neural models outperforms the Majority baseline on unseen domains, but are still worse than the Random baseline. Between both neural models, recurrent networks seem to have an advantage over the Transformer in Discourse Ordering and Text Structuring, whereas the latter approach performs better than the former one in Lexicalization.

\section{Experiment 2: Pipeline vs. End-to-End}

In this experiment, we contrast our pipeline with our end-to-end implementation and state-of-the-art models for RDF-to-text. The models were evaluated in automatic and human evaluations, followed by a qualitative analysis.

\subsection{Approaches}

\paragraph{Pipeline} We evaluated 4 implementations of our pipeline architecture, where the output of the previous step is fed into the next one. We call these implementations \textit{Random}, \textit{Majority}, \textit{GRU} and \textit{Transformer}, where each one has its steps solved by one of the proposed baselines or deep learning implementations. In Random and Majority, the referring expressions were generated by the \textit{OnlyNames} baseline, whereas for GRU and Transformer, NeuralREG was used for the seen entities, \textit{OnlyNames} for the unseen ones and special rules to realize dates and numbers.

\paragraph{End-to-End} We aimed to convert a set of RDF-triples into text using a GRU and a Transformer implementation without explicit intermediate representations in-between.

\subsection{Models for Comparison}

To ground this study with related work, we compared the performance of the proposed approaches with 4 state-of-the-art RDF-to-text models.

\paragraph{Melbourne} is the approach which obtained the highest performance in the automatic evaluation of the WebNLG Challenge. The approach consists of a neural encoder-decoder approach, which encodes a linearized triple set, with predicates split on camel case (e.g. \texttt{floorArea $\rightarrow$ floor area}) and entities represented by general (e.g., \texttt{ENTITY-1}) and named entity recognition (e.g., \texttt{PERSON}) tags, into a template where references are also represented with general tags. The referring expressions are later generated in the template simply by replacing these general tags with an approach similar to \textit{OnlyNames}.

\paragraph{UPF-FORGe} obtained the highest ratings in the human evaluation of the WebNLG challenge, having a performance similar to texts produced by humans. It also follows a pipeline architecture, which maps predicate-argument structures onto sentences by applying a series of rule-based graph-transducers \cite{mille2019}.

\paragraph{\newcite{marcheggiani2018}} proposed a graph convolutional network that directly encodes the input triple set in contrast with previous model that first linearize the input to then decode it into text. 

\paragraph{\newcite{amit2019}} proposed an approach which converts an RDF triple set into text in two steps: text planning, a non-neural method where the input will be ordered and structured, followed by a neural realization step, where the ordered and structured input is converted into text. 

\begin{table*}
\centering
\footnotesize{
\begin{tabular}{l l l l l l l}
\toprule
                         & \textbf{All} & \textbf{Seen} & \textbf{Unseen} & \textbf{All} & \textbf{Seen} & \textbf{Unseen} \\
\midrule
  & \multicolumn{3}{c}{\bf BLEU} & \multicolumn{3}{c}{\bf METEOR} \\
\midrule
Random          & 41.68 & 41.72 & \bf{41.51} & 0.20 & 0.27 & - \\
Majority        & 43.82 & 44.79 & \underline{41.13} & 0.33 & \underline{0.41} & 0.22 \\
GRU             & \underline{50.55} & 55.75 & 38.55 & 0.33 & \bf{0.42} & 0.22 \\
Transformer     & \bf{51.68} & \underline{56.35} & 38.92 & 0.32 & \underline{0.41} & 0.21 \\
E2E GRU         & 33.49 & \bf{57.20} &  6.25 & 0.25 & \underline{0.41} & 0.09 \\
E2E Transformer & 31.88 & 50.79 &  5.88 & 0.25 & 0.39 & 0.09 \\
Melbourne       & 45.13 & 54.52 & 33.27 & \underline{0.37} & \underline{0.41} & \underline{0.33} \\
UPF-FORGe       & 38.65 & 40.88 & 35.70 & \bf{0.39} & 0.40 & \bf{0.37} \\
\cite{marcheggiani2018} & - & 55.90 & - & - & 0.39 & - \\
\cite{amit2019} & 47.40 & - & - & \bf{0.39} & - & - \\
\midrule
\midrule
  & \multicolumn{3}{c}{\bf Fluency} & \multicolumn{3}{c}{\bf Semantic} \\
\midrule
Random          & 4.55$^{E}$  & 4.79$^{D}$  & 4.07$^{D}$  & 4.44$^{D}$  & 4.73$^{D}$   & 3.86$^{C}$ \\
Majority        & 5.00$^{CD}$ & 5.25$^{CD}$ & 4.49$^{CD}$ & 5.02$^{BC}$ & 5.41$^{BC}$  & 4.25$^{BC}$ \\
GRU             & 5.31$^{B}$  & 5.51$^{AB}$ & 4.91$^{BC}$ & 5.21$^{BC}$ & 5.48$^{AB}$  & 4.67$^{B}$ \\
Transformer     & 5.03$^{BC}$ & 5.53$^{AB}$ & 4.05$^{D}$  & 4.87$^{C}$  & 5.49$^{AB}$  & 3.64$^{C}$ \\
E2E GRU         & 4.73$^{DE}$ & 5.40$^{BC}$ & 3.45$^{E}$  & 4.47$^{D}$  & 5.21$^{CD}$  & 3.03$^{D}$ \\
E2E Transformer & 5.02$^{BC}$ & 5.38$^{BC}$ & 4.32$^{CD}$ & 4.70$^{CD}$ & 5.15$^{BCD}$ & 3.81$^{C}$ \\
Melbourne       & 5.04$^{CD}$ & 5.23$^{BC}$ & 4.65$^{CD}$ & 4.94$^{C}$  & 5.33$^{BC}$  & 4.15$^{C}$ \\
UPF-FORGe       & 5.46$^{B}$  & 5.43$^{BC}$ & 5.51$^{AB}$ & 5.31$^{B}$  & 5.35$^{BC}$  & 5.24$^{A}$ \\
\midrule
Original        & 5.76$^{A}$  & 5.82$^{A}$  & 5.63$^{A}$  & 5.74$^{A}$  & 5.80$^{A}$   & 5.63$^{A}$ \\
\bottomrule
\end{tabular}
}
\caption{(1) BLEU and METEOR scores of the models in the automatic evaluation, and (2) Fluency and Semantic obtained in the human evaluation. In the first part, best results are boldfaced and second best ones are underlined. In the second part, ranking was determined by pair-wise Mann-Whitney statistical tests with $p < 0.05$.}
\label{table:results2}
\end{table*}

\subsection{Evaluation}

\paragraph{Automatic Evaluation} We evaluated the textual outputs of each system using the BLEU \cite{papineni2002} and METEOR \cite{lavieAgarwal2007} metrics. The evaluation was done on the entire test data, as well as only in their seen and unseen domains.

\paragraph{Human Evaluation} We conducted a human evaluation, selecting the same 223 samples used in the evaluation of the WebNLG challenge \cite{claire2017b}. For each sample, we used the original texts and the ones generated by our 6 approaches and by the Melbourne and UPF-FORGe reference systems, totaling 2,007 trials. Each trial displayed the triple set and the respective text. The goal of the participants was to rate the trials based on the fluency (i.e., does the text flow in a natural, easy to read manner?) and semantics (i.e., does the text clearly express the data?) of the text in a 1-7 Likert scale.

We recruited 35 raters from Mechanical Turk to participate in the experiment. We first familiarized them with the set-up of the experiment, depicting a trial example in the introduction page accompanied by an explanation. Then each participant had to rate 60 trials, randomly chosen by the system, making sure that each trial was rated at least once.\footnote{The raters had an average age of 32.29 and 40\% were female. 17 participants indicated they were fluent in English, while 18 were native. The experiment, which obtained ethical clearance from the university, took around 20-30 minutes to be completed and each rater received a pay in U.S. dollars for participation.}

\begin{table}
\centering
\footnotesize{
\begin{tabular}{l r r r r r}
\toprule
\multicolumn{6}{c}{\bf Semantic} \\
\midrule
& \textbf{Ord.} & \textbf{Struct.} & \textbf{Txt.} & \textbf{Ovr.} & \textbf{Keep.} \\
\midrule
Random      & 1.00 & 1.00 & 0.43 & 0.05 & 0.41 \\
Majority    & 1.00 & 1.00 & 0.75 & 0.01 & 0.69 \\
GRU         & 0.77 & 0.73 & 0.67 & 0.01 & 0.81 \\
Transformer & 0.75 & 0.69 & 0.68 & 0.08 & 0.80 \\
E2E GRU     & - & - & 0.47 & 0.41 & - \\ 
E2E Trans.  & - & - & 0.39 & 0.53 & - \\ 
Melbourne   & - & - & 0.73 & 0.19 & - \\ 
UPF-FORGe   & - & - & 0.91 & 0.00 & - \\ 
 Original   & - & - & 0.99 & 0.12 & - \\ 
\midrule
\multicolumn{6}{c}{\bf Grammaticality}\\
\midrule
& \textbf{Verb} & \textbf{Det.} & \multicolumn{3}{l}{\textbf{Reference}} \\
\midrule
Random          & 0.95 & 0.91 & 0.89 \\
Majority        & 1.00 & 1.00 & 0.99 \\
GRU             & 1.00 & 0.99 & 0.80 \\
Transformer     & 0.95 & 1.00 & 0.93 \\
E2E GRU         & 0.97 & 1.00 & 0.91 \\
E2E Trans.      & 0.95 & 0.97 & 0.79 \\
Melbourne       & 0.96 & 0.87 & 0.77 \\
UPF-FORGe       & 1.00 & 1.00 & 1.00 \\
Original        & 0.95 & 0.95 & 0.92 \\
\bottomrule
\end{tabular}
}
\caption{Qualitative analysis. The first part shows the percentage of trials that keeps the input predicates over Discourse Ordering (Ord.), Text Structuring (Struct.) and in the final text (Txt.). It also shows the ratio of text trials with more predicates than in the input (Ovr.) and the pipeline texts which keep the decisions of previous steps (Keep.). The second part shows the number of trials without verb, determiner and reference mistakes.}
\label{table:results3}
\end{table}

\paragraph{Qualitative Analysis} To have a better understanding of the positive and negative aspects of each model, we also performed a qualitative analysis, where the second and third authors of this study analyzed the original texts and the ones generated by the previous 8 models for 75 trials extracted from the human evaluation sample for each combination between size and domain of the corpus. The trials were displayed in a similar way to the human evaluation, where the annotators did not know which model produced the text. The only difference was the additional display of the predicted structure by the pipeline approaches (a fake structure was displayed for the other models). Both annotators analyzed grammaticality aspects, like whether the texts had mistakes involving the determiners, verbs and references, and semantic ones, like whether the text followed the predicted order and structure, and whether it verbalizes less or more information than the input triples.\footnote{Inter-annotator agreement for the evaluated aspects ranged from 0.26 (Reference) to 0.93 (Input Triples), with an average Krippendorff $\alpha$ of 0.67.}

\subsection{Results}

Table \ref{table:results2} depicts the results of automatic and human evaluations, whereas Table \ref{table:results3} shows the results of the qualitative analysis.

\paragraph{Automatic Evaluation} In terms of BLEU, our neural pipeline models (GRU and Transformer) outperformed all the reference approaches in all domains, whereas our end-to-end GRU and Random pipeline obtained the best results on seen and unseen domains, respectively.

Regarding METEOR, which includes synonymy matching to score the inputs, reference methods introduced the best scores in all domains. In seen and unseen domains, our neural GRU pipeline and reference approach UPF-FORGe obtained the best results, respectively.

\paragraph{Human Evaluation} In all domains, neural GRU pipeline and UPF-FORGe were rated the highest in fluency by the participants of the evaluation. In seen ones, both our neural pipeline approaches (GRU and Transformer) were rated the best, whereas UPF-FORGe was considered the most fluent approach in unseen domains.

UPF-FORGe was also rated the most semantic approach in all domains, followed by neural GRU and Majority pipeline approaches. For seen domains, similar to the fluency ratings, both our neural pipeline approaches were rated the highest, whereas UPF-FORGe was considered the most semantic approach in unseen domains.

\paragraph{Qualitative Analysis} In general, UPF-FORGe emerges as the system which follows the input the best: 91\% of the evaluated trials verbalized the input triples. Moreover, the annotators did not find any grammatical mistakes in the output of this approach.

When focusing on the neural pipeline approaches, we found that in all the steps up to Text Structuring, the recurrent networks retained more information than the Transformer. However, 68\% of the Transformer's text trials contained all the input triples, against 67\% of the GRU's trials. As in Experiment 1, we see that recurrent networks as GRUs are better in ordering and structuring the discourse, but is outperformed by the Transformer in the Lexicalization step. In terms of fluency, we did not see a substantial difference between both kinds of approaches.

Regarding our end-to-end trials, different from the pipeline ones, less than a half verbalized all the input triples. Moreover, the end-to-end outputs also constantly contained more information than there were in the non-linguistic input. 

\section{Discussion}

This study introduced a systematic comparison between pipeline and end-to-end architectures for data-to-text generation, exploring the role of deep neural networks in the process. In this section we answer the two introduced research questions and additional topics based on our findings.

\paragraph{How well do deep learning methods perform as individual modules in a data-to-text pipeline?} In comparison with Random and Majority baselines, we observed that our deep learning implementations registered a higher performance in the pipeline steps on domains seen during training, but their performance dropped considerably on unseen domains, being lower than the baselines.

In the comparison between our GRU and Transformer, the former seems to be better at ordering and structuring the non-linguistic input, whereas the latter performs better in verbalizing an ordered and structured set of triples. The advantage of GRUs over the Transformer in Discourse Ordering and Text Structuring may be its capacity to implicitly take order information into account. On the other hand, the Transformer could have had difficulties caused by the task's design, where triples and sentences were segmented by tags (e.g. \texttt{<TRIPLE>} and \texttt{<SNT>}), rather than positional embeddings, which suits this model better. In sum, more research needs to be done to set this point.


\begin{figure*}
\footnotesize{
\begin{center}

\begin{tabular}{c c c}
Ace\_Wilder & \underline{\textbf{background}} & ``solo\_singer'' \\
Ace\_Wilder & \textit{\textbf{birthPlace}} & Sweden \\
Ace\_Wilder & \underline{\textbf{birthYear}} & 1982  \\
Ace\_Wilder & \textit{\textbf{occupation}} & Songwriter  \\
&  &   \\
 & $\downarrow$ &   \\
\end{tabular}

{\def\arraystretch{2}\tabcolsep=10pt
\begin{tabular}{l L}
GRU         & Ace Wilder, born in Sweden, performs as Songwriter. \\ \hline
Transformer & Ace Wilder (born in Sweden) was Songwriter. \\
\hline 
E2E GRU     & The test pilot who was born in Willington, who was born in New York, was born in New York and is competing in the competing in the U.S.A. The construction of the city is produced in Mandesh. \\
\hline
E2E Trans. &  Test pilot Elliot See was born in Dallas and died in St. Louis.
\end{tabular}}
\end{center}
\caption{Example of a set of triples from an unseen domain during training (top) and the corresponding texts produced by our pipeline (e.g., GRU and Transformer) and end-to-end approaches (e.g., E2E GRU and E2E Trans.) (bottom). In the top set of triples, predicates seen during training are highlighted in \textit{italic}, whereas the unseen ones are \underline{underlined}.}
\label{fig:example}
}
\end{figure*}

\paragraph{How well does a neural pipeline architecture perform compared to a neural end-to-end one?} Our neural pipeline approaches were superior to the end-to-end ones in most tested circumstances: the former generates more fluent texts which better describes data on all domains of the corpus. The difference is most noticeable for unseen domains, where the performance of end-to-end approaches drops considerably. This shows that end-to-end approaches do not generalize as well as the pipeline ones. In the qualitative analysis, we also found that end-to-end generated texts have the problem of describing non-linguistic representations which are not present in the input, also known as Hallucination \cite{rohrbach2018}. 

The example in Figure \ref{fig:example} shows the advantage of our pipeline approaches in comparison with the end-to-end ones. It depicts the texts produced by the proposed approaches for an unseen set during training of 4 triples, where 2 out of the 4 predicates are present in the WebNLG training set (e.g., \textit{birthPlace} and \textit{occupation}). In this context, the pipeline approaches managed to generate a semantic text based on the two seen predicates, whereas the end-to-end approaches hallucinated texts which have no semantic relation with the non-linguistic input.

\paragraph{Related Work} We compared the proposed approaches with 4 state-of-the-art RDF-to-text systems. Except for \citet{marcheggiani2018}, all the others are not end-to-end approaches, already directing the field to pipeline architectures. UPF-FORGe is a proper pipeline system with several sequential steps, Melbourne first generates a delexicalized template to later realize the referring expressions, and \citet{amit2019} splits the process up into Planning, where ordering and structuring are merged, and Realization.

Besides the approach of \citet{marcheggiani2018}, the ADAPT system, introduced in the WebNLG challenge \cite{claire2017b}, is another full end-to-end approach to the task. It obtained the highest results in the seen part of the WebNLG corpus (BLEU $=60.59$; METEOR $=0.44$). However, the results drastically dropped on the unseen part of the dataset (BLEU $=10.53$; METEOR $=0.19$). Such results correlate with our findings showing the difficult of end-to-end approaches to generalize to new domains.

By obtaining the best results in almost all the evaluated metrics, UPF-FORGe emerges as the best reference system, showing again the advantage of generating text from non-linguistic data in several explicit intermediate representations. However, it is important to observe that the advantage of UPF-FORGe over our pipeline approaches is the fact that it was designed taking the seen and unseen domains of the corpus into account. So in practice, there was no ``unseen" domains for UPF-FORGe. In a fair comparison between this reference system with our neural pipeline approaches in only seen domains, we may see that ours are rated higher in almost all the evaluated metrics.

\paragraph{General Applicability} Besides our findings, the results for other benchmarks, such as \newcite{elder2019}, suggest that pipeline approaches can work well in the context of neural data-to-text generation. Concerning our pipeline approach specifically, although it was designed to convert RDF triples to text, we assume it can be adapted to other domains (and languages) where communicative goals can be linearized and split in units, as in the E2E dataset \cite{novikova2017b}. In future work, we plan to study this in more detail.

\paragraph{Conclusion} In a systematic comparison, we show that adding supervision during the data-to-text process leads to more fluent text that better describes the non-linguistic input data than full end-to-end approaches, confirming the trends in related work in favor of pipeline architectures.

\section*{Acknowledgments}

This work is part of the research program ``Discussion Thread Summarization for Mobile Devices'' (DISCOSUMO) which is financed by the Netherlands Organization for Scientific Research (NWO). We also acknowledge the three reviewers for their insightful comments.

\bibliography{emnlp-ijcnlp-2019}
\bibliographystyle{acl_natbib}

\clearpage
\appendix
\section{Models Set-Up}

\paragraph{General Settings} We used the implementation of \textit{Nematus} \cite{sennrich2017} for both models. We trained each architecture (i.e., GRU and Transformer) three times. For testing, we ensembled the settings which obtained the best results in the development sets in each training execution for GRUs, whereas for the Transformer, we selected the setting which obtained the best result in the respective development set.

Models were trained using stochastic gradient descent with Adam \cite{diederik2015} ($\beta_1=0.9$, $\beta_2=0.98$, $\epsilon=10^{-9}$) for a maximum of 200,000 updates. They were evaluated on the development sets after every 5,000 updates and early stopping was applied with patience 30 based on cross-entropy. Encoder, decoder and softmax embeddings were tied, whereas decoding was performed with beam search of size 5 to predict sequences with length up to 100 tokens. 

\paragraph{GRU Settings} Bidirectional GRUs with attention were used as described in \citet{sennrich2017}. Source and target word embeddings were 300D each, whereas hidden units were 512D.  We applied layer normalization as well as dropout with a probability of 0.1 in both source and target word embeddings and 0.2 for hidden units.

\paragraph{Transformer Settings} Both encoder and decoder consisted of $N = 6$ identical layers. Word embeddings and hidden units were 512D each, whereas the inner dimension of feed-forward sub-layers were 2048D. The multi-head attention sub-layers consisted of 8 heads each. Dropout of 0.1 were applied to the sums of word embeddings and positional encodings, to residual connections, to the feed-forward sub-layers and to attention weights. At training, models had $8000$ warm-up steps and label smoothing of 0.1.   

\paragraph{Word Segmentation} In the lexicalization step of the pipeline and in the end-to-end architecture, \textit{byte-pair encoding} (BPE) \cite{sennrich2016b} was used to segment the tokens of the target template and text, respectively. The model was trained to learn 20,000 merge operations with a threshold of 50 occurrences.

\paragraph{NeuralREG} To generate referring expressions in the pipeline architecture, we used the concatenative-attention version of the NeuralREG algorithm \cite{ferreira2018}. We follow most of the settings in the original paper, except for the number of training epochs, mini-batches, dropout, beam search and early stop of the neural networks, which we respectively set to 60, 80, 0.2, 5 and 10. Another difference is in the input of the model: while NeuralREG in the original paper generates referring expressions based on templates where only the references are delexicalized, here the algorithm generates referring expressions based on a template where verbs and determiners are also delexicalized as previously explained.

\end{document}